%% file: main.tex
\documentclass[conference,letter]{IEEEtran}
\IEEEoverridecommandlockouts
\usepackage{amsmath,amssymb,amsfonts}
\usepackage{algorithmic}
\usepackage{graphicx}
\usepackage{textcomp}
\usepackage[T5]{fontenc}
\usepackage[utf8]{inputenc}
\usepackage{tabularx,ragged2e}
\usepackage[vietnamese,english]{babel}
\usepackage{pifont}
\usepackage{float}
\usepackage{caption}
\usepackage{balance}
\usepackage[export]{adjustbox}
\DeclareCaptionFont{eightpt}{\fontsize{8pt}{10pt}\selectfont #1}
\usepackage[tablename=TABLE]{caption}
\usepackage{mathabx}
\usepackage{multirow}
\usepackage{graphicx}

\usepackage[skip=3pt]{caption}

\usepackage[backend=biber,
sorting=none,
doi=false,
isbn=false,
url=true,
maxcitenames=2,
mincitenames=1,
style=ieee,]{biblatex}

\begin{document}
%Paper title
\title{Investigating Monolingual and Multilingual BERT Models for Vietnamese Aspect Category Detection}
%Paper Authors
\author{\IEEEauthorblockN{1\textsuperscript{st} Dang Van Thin} \IEEEauthorblockA{\textit{\textit{Multimedia Communications Laboratory (MMLab)}}\\\textit{University of Information Technology}\\\textit{Vietnam National University, Ho Chi Minh City}\\ Vietnam\\Email: thindv@uit.edu.vn}\\
 \IEEEauthorblockN{3\textsuperscript{rd} Vu Xuan Hoang} \IEEEauthorblockA{\textit{Department of Computer Science}\\\textit{University of Information Technology}\\\textit{Vietnam National University, Ho Chi Minh City}\\ Vietnam\\Email: 19522531@gm.uit.edu.vn} 
 \and \IEEEauthorblockN{2\textsuperscript{nd} Lac Si Le}\IEEEauthorblockA{\textit{Department of Software Engineering}\\\textit{University of Information Technology}\\\textit{Vietnam National University, Ho Chi Minh City}\\ Vietnam\\ Email: 17520669@gm.uit.edu.vn}\\ \IEEEauthorblockN{4\textsuperscript{th} Ngan Luu-Thuy Nguyen} \IEEEauthorblockA{\textit{Faculty of Information Science and Engineering}\\\textit{University of Information Technology}\\\textit{Vietnam National University, Ho Chi Minh City}\\ Vietnam\\ Email: ngannlt@uit.edu.vn} }

\maketitle
\IEEEoverridecommandlockouts
\IEEEpubidadjcol

%Paper Abstract
\begin{abstract}
Aspect category detection (ACD) is one of the challenging tasks in the Aspect-based sentiment Analysis problem. The purpose of this task is to identify the aspect categories mentioned in user-generated reviews from a set of pre-defined categories. In this paper, we investigate the performance of various monolingual pre-trained language models compared with multilingual models on the Vietnamese aspect category detection problem. We conduct the experiments on two benchmark datasets for the restaurant and hotel domain. The experimental results demonstrated the effectiveness of the monolingual PhoBERT model than others on two datasets. We also evaluate the performance of the multilingual model based on the combination of whole SemEval-2016 datasets in other languages with the Vietnamese dataset. To the best of our knowledge, our research study is the first attempt at performing various available pre-trained language models on aspect category detection task and utilize the datasets from other languages based on multilingual models.

\end{abstract}

%Paper Ketwords

\input{sections/1-Introduction.tex}
\input{sections/2-Relatedwork.tex}
\input{sections/3-Architecture.tex}

\input{sections/4-Experiments.tex}

\input{sections/6-Conclusion.tex}

\section*{Acknowledgment}
We thank the anonymous reviewers for their valuable comment for this manuscript. We would greatly appreciate any feedback on this research, please send an email to the first author for questions about this paper.
\printbibliography{}

\end{document}

%% file: sections/1-Introduction.tex
\section{Introduction} \label{introduction}
With the growth of e-commerce sites, user-generated reviews have a great influence on the success of the products or services. Besides, these reviews also are freely valuable information for both the users and the producers in term of development of products.  Aspect category Detection is one of the main tasks of Aspect-based Sentiment Analysis problem which aims to identify the aspect categories mentioned in the review. The formula of this task is presented as follows: Give a user-generated review and the list of pre-defined categories for the specific domain, the purpose of this task is to detect the explicit or implitcit the aspect categories in the review. For example, given a review for the restaurant domain, ``The sushi seemed pretty fresh and was adequately proportioned.'', there are two aspect categories belongs to the review as Food\#Quality and Food\#Style\_Options.

Up to now, there are several studies have been carried out Aspect Category Detection task on different languages such as English, German, etc, \cite{10.5555/2887007.2887066, schouten2017supervised, ghadery2019mncn, ghadery2018, 9092619, movahedi2019aspect} . Most of studies have been proposed several models to address this task based on the supervised learning model, deep learning model and BERT-based model. For the low-resource language such as Vietnamese, there are a few studies on Aspect Category Detection task such as \citeauthor{thin_nics_2018}\cite{thin_nics_2018}, \citeauthor{thuy2018cross}\cite{thuy2018cross} . However, they only focus on designing the architecture based on the supervised learning and deep learning methods to solve this task. Their results are quite impressive, however, these methods are old-fashioned compared with state of the art technique in the development of the Natural Language Processing (NLP) field. Recently, BERT architecture \cite{devlin2018bert} has shown the effectiveness on various tasks in the NLP field, including ACD task. Besides, there are many variants of BERT models which are published for Vietnamese language and multi-lingual model as cross-lingual language model (XLM-R) \cite{conneau_XLMR} or multilingual BERT \cite{devlin2018bert}. These models are pre-trained on Wikipedia and Common crawl data for different languages, including Vietnamese. For those reasons, in this paper, we investigate the performance of BERT-based model for ACD task in Vietnamese based on the various mono-lingual models compared with the multi-lingual models. In addition, we also evaluate the performance of the combination Vietnamese with other datasets and cross-lingual based on the multilingual XLM-R architecture. 

The rest of the paper is structured as follows. Section 2 is an overview of related work about aspect category detection. Section 3 presents the model which is used for experiments in this paper. Then, we present the whole experiments including datasets, model settings and our results in Section 4. Section 5 presents the conclusion and gives some future works.  

%% file: sections/2-Relatedwork.tex
\section{Related Work}
\label{relatedworks}
There are a number of studies works focus on design the algorithm and architecture to address the Aspect Category Detection task. This task is one of the main tasks of the Aspect-based sentiment analysis challenge and was official introduced to the community in  SemEval workshop 2014 \cite{pontiki-etal-2014-semeval}. Most of the previous studies on this task are based on supervised machine learning techniques, some of the works also present the unsupervised system. Among the classic classification algorithms, the Support Vector Machine (SVM) is a powerful algorithm most selected for this task \cite{kiritchenko-etal-2014-nrc, xenos-etal-2016-aueb, ganu2009beyond, thin_journal_2019, thuy2018cross} . The authors used the one-vs-all or multiple binary SVM classifiers combine with various handcraft features.

In recent years, with the development of deep learning model, there are various appoach proposed to solve the ACD task. First, Zhou et al., \cite{10.5555/2887007.2887066} presented a representation learning
approach for the ACD task based on the semi-supervised word embedding along with the hybrid feature extraction through neural networks stacked on the word vectors. Finally, a logistic regression classifier is trained to predict the aspect of category using hybrid features. While, \citeauthor{van2018deep}\cite{van2018deep} and \citeauthor{9092619}\cite{9092619} presented a Convolutional Neural Network architecture for the ACD task for the Vietnamese and Indonesian language. \citeauthor{xue-etal-2017-mtna}\cite{xue-etal-2017-mtna} present a neural networks based on the BiLSTM layer combined with CNN layer for the aspect category and aspect term tasks. After that, \citeauthor{sajad_movahedi}\cite{sajad_movahedi} presented the architecture of Topic-Attention Network which is based on attention mechanism to explore the relation of words different different aspect categories. Their experimental results on two SemEval datasets in the restaurant domain shown the effectiveness of proposed architecture. Recently, the BERT \cite{devlin2018bert} architecture is marked as the beginning of a new are in the NLP field. \citeauthor{Ramezani2020AspectCD} \cite{Ramezani2020AspectCD} presented a study on contextual representations of the BERT model to better extract useful features for the ACD task. 

For the unsupervised method, the first method of this approach that applied association rule mining on co-occurrence frequency data obtained from a corpus to find these aspect categories is presented by \citeauthor{schouten2017supervised}\cite{schouten2017supervised} and \citeauthor{ghadery2018} \cite{ghadery2018}. However, the functional drawback of this unsupervised approach is that it involves the tuning of several parameters. For the multilingual ACD task, \cite{ghadery2019mncn} \cite{ghadery2019mncn} proposed a multilingual method based on Deep CNN combined with multilingual embedding on multiple languages at the same time. These results have shown the effectiveness of the multilingual approach. The work of \cite{thuy2018cross} also presented a study on cross-language ACD task, however, they just translated the English dataset to the Vietnamese and combined them with their target dataset (in Vietnamese) to increase the training set size for SVM model. 

%% file: sections/3-Architecture.tex
\section{Model} \label{Model}

\begin{figure*}[htbp]
\centering
\includegraphics[width=0.5\textwidth]{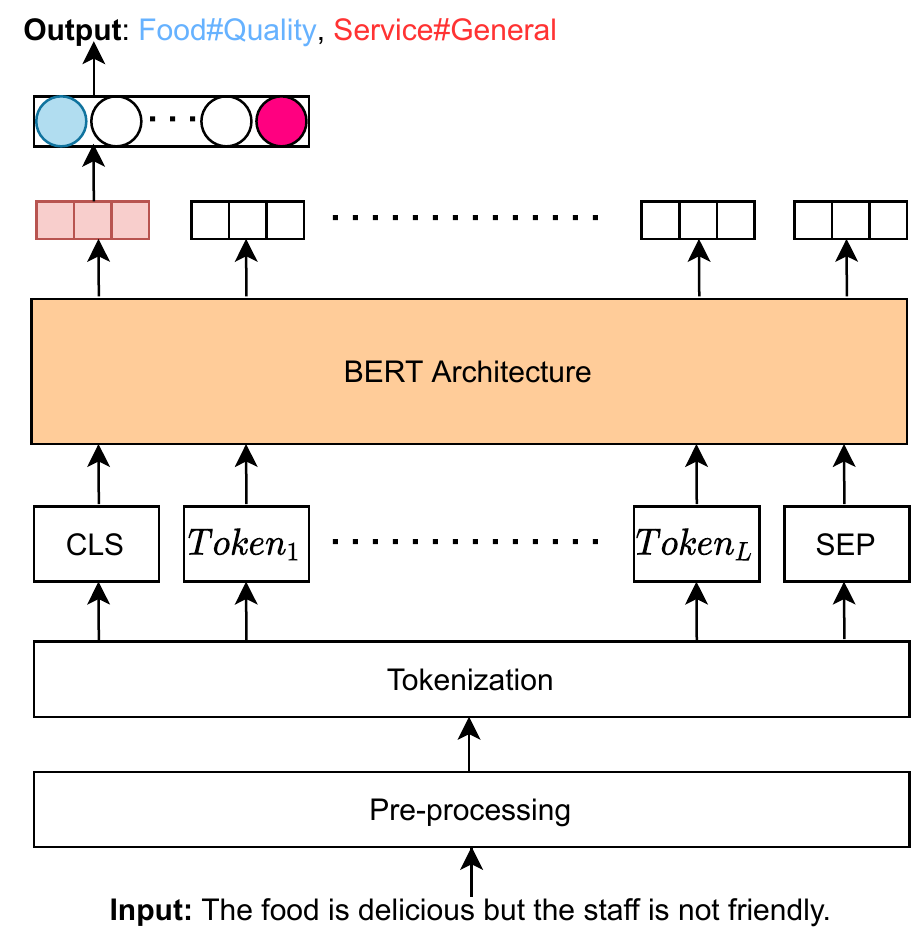}
\caption{The overall architecture for the aspect category detection on users review.}
\label{fig1}
\end{figure*}

\subsection{Model Description}
In this section, we introduce the design of our model based on BERT-based architecture for the Aspect Category Detection task. The overall architecture is depicted in Figure \ref{fig1}. In general, this task aims to identify the categories mentioned in the review. Formally, given a review as the input with L words $S = \{s_1, s_2, ..., s_L\}$ where $s_i$ is the \textit{i-th} word in the review, and list of \textit{N} pre-defined aspect categories $A = \{a_1, a_2, ..., a_N\}$. We must identify a list of aspect categories mentioned in the input review. For examples, given a review in the restaurant domain ``The food is delicious but the staff is not friendly'', the output should be as ``Food\#Quality'' and ``Service\#General''.    

As shown in Figure \ref{fig1}, the text input is processed in a pre-processing component before putting into the tokenizer component. These steps in the pre-processing component will be depended on the implementation of the pre-trained model. We refer carefully to the way the BERT model create a pre-training dataset to decide the steps in the pre-processing component. For example, the BERT model use the word segmentation technique for the training text, we also apply this technique for our review dataset. Next, the tokenizer is responsible for splitting the pre-processed sentence into tokens and mapping these tokens to their index in the tokenizer vocabulary of BERT architecture. This component also adds special tokens to the start [CLS] and the end of each input sentence [SEP] automatically. Because the reviews in the dataset are of varying length, therefore we must use padding to make all the inputs have the same length for each dataset. In our case, we can use the maximum sequence length to pad the dataset. Then the list of token's ids with attention mark is feed into the pre-trained BERT model to extract the hidden state of tokens. For the pre-trained BERT architecture, we freeze all the layers of the BERT model and only update the weights of the prediction layer during model training. Based on the work of Devlin et al.,\cite{devlin2018bert} , we only use the [CLS] representation to fed into an output layer with sigmoid activation function $\sigma$ for calculating the probability of label. The output range of this function is between zero and one; if the probability of a category is above a cut-off threshold then this category is assigned to the input sentence. The threshold can be optimized by using a grid search technique on the validation set. We use the binary cross-entropy as the loss function as follow:

\begin{IEEEeqnarray}{rCl}
L = - \sum_{c=1}^{C} y_c \cdot \log (\sigma (\hat{y}_c)) + (1 - y_c) \cdot \log(1 - \sigma(\hat{y}_c)) 
\end{IEEEeqnarray}

where C is the number of classes, $\sigma$ is a sigmoid activation function.

\subsection{BERT Models}
In this section, we introduce the various pre-trained BERT models for Vietnamese and multilingual language which are used in this paper.

% \usepackage{graphicx}
\begin{table*}[t]
\centering
\caption{Comparison of various pre-trained language models
for Vietnamese.}
\label{table_vbert}
\resizebox{0.7\textwidth}{!}{%
\begin{tabular}{lcccc}
\hline
\textbf{Information} &
  \multicolumn{1}{l}{\textbf{PhoBERT}} &
  \multicolumn{1}{l}{\textbf{viBert4news}} &
  \multicolumn{1}{l}{\textbf{viBert\_FPT}} &
  \multicolumn{1}{l}{\textbf{vELECTRA\_FPT}} \\ \hline
Data Domain           & News+Wiki & News     & News    & News    \\
Data Size             & 20G         & 20GB     & 10GB    & 60GB    \\
Tokenization          & Subword     & Syllable & Subword & Subword \\
Vocabulary size       & 64K         & 62K      & 32K     & 32K     \\
Word segmentation & True        & False    & False   & False   \\ \hline
\end{tabular}%
}
\end{table*}

\subsubsection{Vietnamese BERT models}

\begin{itemize}
    \item \textbf{viBert\_FPT}\cite{the2020improving}: This pre-trained model trained based on the standard multi-lingual BERT (mBERT) \footnote{https://github.com/google-research/bert/blob/master/multilingual.md} package with 12 blocks of transformers. The list of vocabularies of this model is modified from original vocabularies of mBERT This model is trained on a 10GB corpus forthe News domain based on the subword tokenizer. 
    \item \textbf{vELECTRA\_FPT} \cite{the2020improving}: ELECTRA can be used to pre-train transformer networks using relatively little compute for self-supervised language representation learning \cite{clark2020electra}. To train this model for Vietnamese, the authors collect two sources with a 60GB text including the NewsCorpus\footnote{https://github.com/binhvq/news-corpus} and the OscarCorpus\footnote{https://oscar-corpus.com/}.
    
    \item \textbf{viBert4news}\footnote{https://github.com/bino282/bert4news/}: This pre-trained model is also trained based on the BERT architecture on 20GB dataset by using word sentence piece with basic bert tokenization and same config with BERT base without lowercasing. This model is used to fine-turning to NLP problems in Vietnamese such as Word Segmentation, Named Entity Recognition\footnote{https://github.com/bino282/ViNLP}.
    
    \item \textbf{PhoBERT} \cite{nguyen2020phobert}: PhoBERT has two versions using the same architectures of BERT, however, these models used the RoBERTa \cite{liu2019roberta} to optimize the BERT pre-training procedure. They applied the fastBPE tokenizer to segment word-segmented sentences with subword units. Their experimental results demonstrated the effectiveness of PhoBERT for four downstream Vietnamese NLP tasks.
    
\end{itemize}
Table \ref{table_vbert} present the comparison of four pre-trained BERT language models for the Vietnamese language, including data domain, vocabulary size, etc.

\subsubsection{Multilingual BERT models}

\begin{itemize}
    \item \textbf{mBERT} \cite{devlin2018bert}: Multilingual BERT (mBERT) provides cross-lingual representations for 104 languages, including Vietnamese. In this paper, we evaluate the quality of mBERT representation on Vietnamese language for the Aspect Detection Task. Besides, we also explore how mBERT performs on cross-lingual between Vietnamese and other languages.
    
    \item \textbf{XLM-R} \cite{conneau_XLMR}: This is a transformer-based multilingual masked language model which trained on text in 100 languages. This model outperforms mBERT on various cross-lingual tasks on low-resource languages. Their experimental results also demonstrated the performance of XLM-R has competitive with SOTA monolingual models.
    
    \item \textbf{mDistilBert} \cite{sanh2019distilbert}: This DistilBERT model is distilled from the multilingual BERT model \cite{devlin2018bert} with the small size than original but still retains the capabilities of BERT model.
\end{itemize}

%% file: sections/4-Experiments.tex
\section{Experiments} \label{Experiments}
\subsection{Datasets}
In this paper, we used two Vietnamese benchmark datasets \cite{thin_tallip} and SemEval-2016 task 5 datasets \cite{pontiki-etal-2016-semeval} in the restaurant and hotel domain at the sentence-level in our experiments. For the SemEval-2016 datasets, we use the whole datasets in the restaurant domain and we discarded sentences which do not contain any aspect categories. We aim to use these datasets combined with Vietnamese training dataset based on multilingual pre-trained BERT models. We want to evaluate the effectiveness of the multilingual models when adding more datasets from other languages on the Vietnamese testing set.

\subsection{Experiment Settings}
In this section, we present the hyper-parameters of model and information of various pre-trained BERT models which are used in this paper. For pre-processing component, we re-implemented these steps as research works \cite{van2018transformation, van2019multi} on Vietnamese ABSA. In addition, each pre-trained model has the specific pre-processing steps for the text data, therefore, we read and add carefully these pre-processing steps on the text input. For example, we use VnCoreNLP's word segmenter \cite{vu-etal-2018-vncorenlp} to pre-process the reviews before using the PhoBERT model as its recommendation. For other languages, we only convert text to lowercase, remove multiple spaces, remove punctuation and replace numbers as a special word. We use the Rectified Adam optimizer with learning rate as 2e-5, warm up proportion as 0.1. We set the batch size as 32, the maximum number of epochs as 50 with the early stopping strategy. The threshold is set as 0.5. For the monolingual pre-trained BERT model, we use the $PhoBERT_{base}$\footnote{https://github.com/VinAIResearch/PhoBERT}, viBERT\_FPT\footnote{https://github.com/fpt-corp/viBERT}, vELECTRA\_FPT\footnote{https://github.com/fpt-corp/vELECTRA}, vibert4news\footnote{https://github.com/bino282/bert4news/} as their recommendation. For the multilingual BERT model, we use the available pre-trained models on Huggingface\footnote{https://huggingface.co/transformers/pretrained\_models.html} including mBERT\footnote{bert-base-multilingual-cased}, mDistilbert\footnote{distilbert-base-multilingual-cased}, XLM-R\footnote{xlm-roberta-base}. As the evaluation metric, we use micro-averaged F1-score, Precision, and Recall as previous studies \cite{pontiki-etal-2016-semeval}. 

\subsection{Experimental Results}

Table \ref{table4} shows the comparison results for different pre-trained language models for the restaurant domain, while Table \ref{table5} presents the experimental results for the hotel domain. It is very interesting that PhoBERT model achieved results superior to the other models in terms of F1-score for two domains, which indicates the effectiveness of PhoBERT in performing aspect category detection in the Vietnamese language. Comparing the results of PhoBERT against other Vietnamese monolingual models, we observe that PhoBERT with word-level/subword tokenization achieved the best results on two test sets. One of the reasons for the good performance of the PhoBERT model is trained on word-level Vietnamese corpus where Vietnamese words can consist of one (single words) or more syllables (compound words). For other Vietnamese BERT model, vElectra\_FPT gained a slightly better than ViBERT4news and viBERT\_FPT in term of F1-score. Among three multi-lingual models, XLM-R is still the best model for the Vietnamese language, furthermore, this model gives a result higher than three monolingual language pre-trained models (vBERT4news, viBERT\_FPT, vElectra\_FPT) in term of F1-score for the restaurant domain. While the results of models are similar in the hotel domain. All experimental results demonstrated the effectiveness of monolingual pre-trained language models PhoBERT for Vietnamese tasks. These results demonstrated that the tasks as text classification can achieve better performance with the pre-trained word-level model than syllable-level model for the Vietnamese language.

\begin{table}[t]
\centering
\caption{The experimental results of various mono-lingual and multi-lingual pre-trained BERT models on Vietnamese aspect category detection task for the restaurant domain.}
\label{table4}
\begin{tabular}{l|lccc}
\hline
\textbf{Types} & \textbf{Models} & \textbf{Precision} & \textbf{Recall} & \textbf{F1-score} \\ \hline
\multirow{3}{*}{Multi-lingual} & mBERT & 81.39 & 76.34 & 78.78 \\ \cline{2-5} 
 & mDistilBert & 80.35 & 76.07 & 78.16 \\ \cline{2-5} 
 & XLM-R & 82.98 & 81.40 & 82.18 \\ \hline
\multirow{4}{*}{Mono-lingual} & viBert4news & 79.26 & 77.48 & 78.36 \\ \cline{2-5} 
 & viBert\_FPT & 80.65 & 79.12 & 79.88 \\ \cline{2-5} 
 & vELECTRA\_FPT & 83.08 & 79.54 & 81.27 \\ \cline{2-5} 
 & PhoBERT & \textbf{85.60} & \textbf{87.49} & \textbf{86.53} \\ \hline
\end{tabular}
\end{table}

\begin{table}[t]
\centering
\caption{The experimental results of various mono-lingual and multi-lingual pre-trained BERT models on Vietnamese aspect category detection task for the hotel domain.}
\label{table5}
\begin{tabular}{l|lccc}
\hline
\textbf{Types} & \textbf{Models} & \textbf{Precision} & \textbf{Recall} & \textbf{F1-score} \\ \hline
\multirow{3}{*}{Multi-lingual} & mBERT & 77.93 & 76.26 & 77.09 \\ \cline{2-5} 
 & mDistilBert & 78.59 & 74.97 & 76.73 \\ \cline{2-5} 
 & XLM-R & 78.86 & 76.56 & 77.70 \\ \hline
\multirow{4}{*}{Mono-lingual} & viBert4news & 79.39 & 74.83 & 77.04 \\ \cline{2-5} 
 & viBert\_FPT & 81.14 & 74.54 & 77.70 \\ \cline{2-5} 
 & vELECTRA\_FPT & 79.82 & 76.07 & 77.90 \\ \cline{2-5} 
 & PhoBERT & \textbf{81.49} & \textbf{76.96} & \textbf{79.16} \\ \hline
\end{tabular}
\end{table}

\begin{table}[t]
\centering
\caption{The experimental results of multi-lingual pre-trained BERT models on Vietnamese test dataset by using the Vietnamese combined with whole SemEval-2016 training \cite{pontiki-etal-2016-semeval} datasets in other languages as training set.}
\label{table55}
\begin{tabular}{l|lccc}
\hline
\textbf{Domain} & \textbf{Models} & \textbf{Precision} & \textbf{Recall} & \textbf{F1-score} \\ \hline
\multirow{3}{*}{Restaurant} & mBERT & 79.22 & 79.92 & 79.57 \\ \cline{2-5} 
 & mDistilBert & 80.76 & 76.54 & 78.59 \\ \cline{2-5} 
 & XLM-R & \textbf{83.48} & \textbf{83.99} & \textbf{83.73} \\ \hline
\multirow{3}{*}{Hotel} & mBERT & 77.93 & 76.38 & 77.15 \\ \cline{2-5} 
 & mDistilBert & \textbf{79.85} & 74.39 & 77.02 \\ \cline{2-5} 
 & XLM-R & 79.25 & \textbf{76.75} & \textbf{77.98} \\ \hline
\end{tabular}
\end{table}

\begin{table}[t]
\caption{The experimental results on the Vietnamese test dataset based on the pre-trained XLM-R model for the restaurant domain. Vi-EN means that the training dataset inlcudes the Vietnamese combined with English dataset.}
\centering
\label{table666}
\begin{tabular}{l|c|c|c}
\textbf{Dataset} & \multicolumn{1}{l|}{\textbf{Precision}} & \multicolumn{1}{l|}{\textbf{Recall}} & \multicolumn{1}{l}{\textbf{F1-score}} \\ \hline
Vi+En & 84.43 & 84.37 & \textbf{84.40} \\ \hline
Vi+Fr & 83.17 & 82.73 & 82.95 \\ \hline
Vi+Du & 81.86 & 84.10 & 82.96 \\ \hline
Vi+Tu & 83.69 & 83.34 & 83.51 \\ \hline
Vi+Ru & \textbf{83.72} & 82.77 & 83.24 \\ \hline
Vi+Sp & 81.82 & \textbf{86.31} & 84.01 \\ \hline
En+Sp+Fr+Du+Tu+Ru & 63.32 & 56.60 & 59.77 \\ \hline
\end{tabular}
\end{table}

The second research in this paper is to study the effectiveness of multilingual models when combining data from other languages to increase the samples in the training set. Table \ref{table55} shows the results of three multilingual models on the combination of Vietnamese dataset with whole SemEval-2016 dataset for each domain. We observe that most of the models achieve a higher performance than those models where is trained only on the Vietnamese training set (see Table \ref{table4} and Table \ref{table5}). We also investigate the performance of the combination of Vietnamese with each language based on the XLM-R model for the restaurant domain. Table \ref{table666} presents the results of XLM-R model on the combination Vietnamese with a single language (e.g. English or French, etc.,) as the training set and Vietnamese as the testing set. We can see that the combination of English with Vietnamese as the training set gives the best F1-score than others. The last row in Table \ref{table666} shows the result of XML-R model on the combination of all languages without Vietnamese as the training set and Vietnamese as the testing set. The result shows that training multilingual models on other languages without target dataset do not give the satisfying performance on the target testing set. Based on the experimental results in Table \ref{table666}, it is worth noting that the combination of the source language with target language can affect the performance of models on the testing set, for example, the combination of English with Vietnamese training set gains the better results in terms of F1-score than the combination of all available languages in our case.

%% file: sections/6-Conclusion.tex
\section{Conclusion And Future Work} \label{conclusion}

In this paper, we investigated the effectiveness of mono-lingual compared with multi-lingual pre-trained BERT model for Vietnamese aspect category detection problem. Empirical results prove the effectiveness of PhoBERT compared to several models, including the XLM-R, mBERT model and another version of BERT model for Vietnamese languages. In addition, We also conducted extensive experiments to demonstrate the effectiveness of mono-lingual pre-trained BERT language models compared against multi-lingual models. Our empirical results on multilingual datasets demonstrate that combining dataset from other languages using multilingual models can improve the performance on the target language. For future works, we intend to expand this approach by analysing more experimental results and extend for other tasks in Vietnamese languages.